\documentclass[runningheads]{llncs}
\usepackage{graphicx}
\usepackage{multirow}

\usepackage{xcolor}         
\usepackage{pifont}

\newcommand{\ignore}[1]{}

\usepackage{amsfonts}
\usepackage{latexsym}

\begin{document}
%

\title{SEIHAI: A Sample-efficient Hierarchical AI for the MineRL Competition}
%
%
\author{Hangyu Mao$^\dagger$\inst{1}\and
Chao Wang$^\dagger$\inst{1} \and
Xiaotian Hao$^\dagger$\inst{2} \and
Yihuan Mao$^\dagger$\inst{3} \and
Yiming Lu$^\dagger$\inst{3} \and
Chengjie Wu$^\dagger$\inst{3} \and
Jianye Hao$^\ddagger$\inst{1,2} \and
Dong Li\inst{1} \and
Pingzhong Tang\inst{3}}
\authorrunning{H. Mao, C. Wang, X. Hao, Y. Mao, Y. Lu, C. Wu, J. Hao et al.}
%
\institute{Huawei Noah's Ark Lab, China \\ \email{\{maohangyu1,wangchao358,haojianye,lidong106\}@huawei.com} \and
Tianjin University, China \\ \email{\{xiaotianhao,jianye.hao\}@tju.edu.cn} \and
IIIS, Tsinghua University, China \\
\email{\{maoyh20,luym19,wucj19\}@mails.tsinghua.edu.cn}, \email{kenshin@tsinghua.edu.cn}
}

\maketitle              
\def\thefootnote{$^\dagger$}\footnotetext{These authors contribute equally to this work.}\def\thefootnote{\arabic{footnote}}
\def\thefootnote{$^\ddagger$}\footnotetext{Jianye Hao is the corresponding author.}\def\thefootnote{\arabic{footnote}}

\begin{abstract}
The MineRL competition is designed for the development of reinforcement learning and imitation learning algorithms that can efficiently leverage human demonstrations to drastically reduce the number of environment interactions needed to solve the complex \emph{ObtainDiamond} task with sparse rewards. To address the challenge, in this paper, we present \textbf{SEIHAI}, a \textbf{S}ample-\textbf{e}ff\textbf{i}cient \textbf{H}ierarchical \textbf{AI}, that fully takes advantage of the human demonstrations and the task structure. Specifically, we split the task into several sequentially dependent subtasks, and train a suitable agent for each subtask using reinforcement learning and imitation learning. We further design a scheduler to select different agents for different subtasks automatically. SEIHAI takes the first place in the preliminary and final of the NeurIPS-2020 MineRL competition. 

\keywords{MineRL Competition \and Reinforcement Learning \and Imitation Learning \and Sample Efficiency}
\end{abstract}

\section{Introduction}
Reinforcement learning has achieved tremendous breakthroughs in games \cite{mnih2015human}, robotic manipulations \cite{gu2017deep}, network configurations \cite{mao2019modelling,mao2020neighborhood}, and multi-agent systems \cite{mao2017accnet,mao2020learningAgent,mao2020learningMultiAgent}. However, the state-of-the-art reinforcement learning algorithms usually require a large number of environment interactions, which can be expensive or even infeasible in real-world applications. In addition, most methods can hardly be applied directly to tasks with very sparse rewards.  

Recently, the MineRL competition \cite{guss2019minerldataset,guss2019neurips,milani2020retrospective,guss2021towards,guss2021minerl,guss2019minerl} has been introduced in order to promote researches in sample-efficient reinforcement learning and imitation learning that leverage human demonstrations to solve the complex \emph{ObtainDiamond} task in the Minecraft game. Specifically, an agent starts off from a random position on a randomly-generated Minecraft map with the goal to obtain a Diamond, which can only be accomplished by mining materials and crafting necessary tools. Moreover, mining or crafting some items requires crafting other prerequisite items. For instance, a wooden pickaxe is needed for collecting stones, which is further required by other more advanced tools. Besides, an agent would be rewarded if and only if it successfully obtains 
the right item at the right time. In other word, the environment is with sparse rewards.
In summary, the \emph{ObtainDiamond} task requires the completion of a series of sequentially dependent subtasks with sparse rewards, which usually takes thousands of steps. The randomness of generated worlds, the visual observations with low signal-to-noise ratios, and the temporal dependence of underlying subtasks and the sparse-reward setting constitute the challenges for efficient learning.

In this paper, we leverage the hierarchical dependency property of different items, and propose a hierarchical solver to address the challenge of the MineRL competition. Specifically, we split the \emph{ObtainDiamond} task into several subtasks according to the natural sequential dependency among the items, and train a suitable reinforcement learning or imitation learning agent to tackle each subtask. To further alleviate the sparse-reward problem and improve the sample efficiency, the task structure knowledge and the human demonstrations are employed. Moreover, to make our methods fully automatic and generalizable to other domains \footnote{The competition requires that the submitted methods should not include any meta-actions or rule-based heuristics.}, we design a scheduler to select different agent in different game states. The proposed hierarchical agent SEIHAI is optimized based on limited human demonstrations and environment interactions as the competition required.\footnote{Specifically, over 60 million frames of human demonstrations and 8 million online interactions with the environment.} Our final model wins the first place in both the preliminary and final of the MineRL competition at NeurIPS-2020, among 90+ teams and around 500 submissions.
The contributions are summarized as follows.

1) We proposed a fully automatic hierarchical method to address the \emph{ObtainDiamond} task with sparse rewards and limited human demonstrations in the MineRL competition.

2) We introduced a few learning-based methods to extract critical actions from the large continuous and domain-agnostic action spaces, which enhances the sample efficiency of our methods.

3) We evaluated our methods in the MineRL competition, and won the first place in the preliminary and final of NeurIPS-2020 MineRL competition. The results demonstrate the effectiveness of our methods.

\section{Background}
\subsection{The MineRL Competition}
The MineRL competition is designed to promote the development of sample-efficient, domain-agnostic, and robust algorithms for solving complex tasks with sparse rewards using human demonstrations \cite{guss2021minerl}. The competition consists of the following challenges.

\textbf{Limited Human Demonstrations and Environment Interactions.} The competition provides 60 million frames of human demonstrations \cite{guss2019minerldataset}, but only 211 trajectories are collected in the \emph{ObtainDiamond} environment\footnote{Most human demonstrations are scattered among several small environments, like \emph{Navigation}, \emph{Treechop}.}. Besides, participants are restricted to use at most 8 million frames of environment interactions. Therefore, it is necessary to use imitation learning techniques to fully leverage the demonstrations.

\textbf{Domain Agnosticism.} The original states and discrete actions have been obfuscated into continuous feature vectors by an unknown autoencoder. This prevents participants from using domain-specific actions or rule-based heuristics.
\ignore{
\textbf{Reproducibility and Robustness.} After submission, the code will be retrained from scratch in a re-rendered environment with a corresponding private dataset in the testing. The rendering techniques include but not limited to texture remapping and action remapping. Thus, it is critical to develop robust methods agnostic to these changes, and can reproduce satisfactory performances steadily.}

\subsection{The ObtainDiamond Task}
The goal of the MineRL competition is to address the \emph{ObtainDiamond} task. This task is featured with the following challenges:  

\begin{figure}[t!]
    \includegraphics[width=\textwidth]{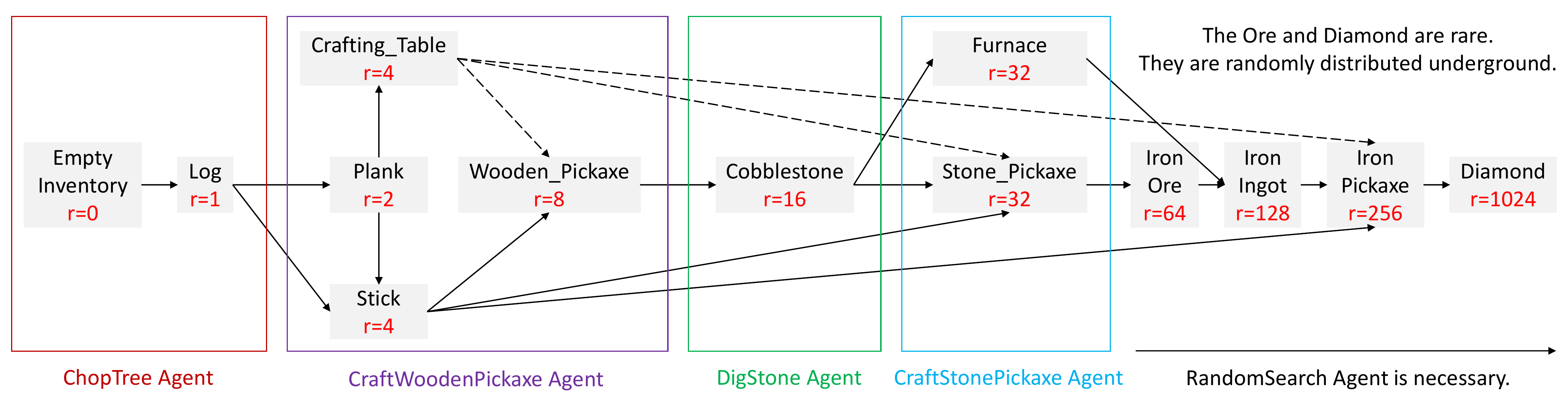}
    \caption{The background about the MineRL competition: the grey rectangles represent different items, e.g, Log, Plank, Crafting\_Table, Stick, Wooden\_Pickaxe, and so on; the arrows demonstrate the typical item hierarchy for obtaining a Diamond; the red texts denote the reward when the agent mines or crafts an item for the first time. We split the \emph{ObtainDiamond} task into five subtasks according to the natural hierarchy among items, and train a suitable agent for each subtask.} 
    \label{fig:ItemHierarchy}
\end{figure}

\textbf{Item Dependency.} In order to obtain a Diamond, the agent needs to craft or obtain different items in a certain order. The typical item dependency (or item hierarchy) for obtaining a Diamond is shown in Figure \ref{fig:ItemHierarchy}. To finish the task, the agent must follow the item hierarchy, otherwise it will not get any rewards. For example, it is impossible for the agent to dig Cobblestones before crafting the Wooden\_Pickaxe. 

\textbf{Sparse Reward.} The agent receives a reward \emph{the first time} it crafts or obtains an item, and the different reward value are depicted in Figure \ref{fig:ItemHierarchy}. Note that obtaining the same item several times does not incur additional rewards, although it is necessary for the agent to collect enough base items in order to craft all the tools it needs. 

\textbf{Long Episode Length.} As can be seen from Figure \ref{fig:ItemHierarchy}, mining or crafting each item takes lots of steps. According to the human demonstrations, it usually takes more than thousands of steps, for a skilled human to obtain the Diamond. As a result, it aggravates the sparse-reward situation.  

\subsection{The ObtainDiamond MDP}
The \emph{ObtainDiamond} environment can be formulated as a Markov Decision Process (MDP), which can be formally defined by the tuple $\langle S, A, T, R, \gamma \rangle$, where $S$ is the set of possible states $s$; $A$ represents the set of possible actions $a$; $T(s'|s,a): S \times A \times S \rightarrow [0,1]$ denotes the state transition function (or the environment dynamics); $R(s,a): S \times A \rightarrow \mathbb{R}$ is the reward function; $\gamma \in [0,1]$ is the discount factor.

We use $s_t$, $a_t$ and $r_t=R(s_t,a_t)$ to denote the state, action and reward at time step $t$, respectively. In the \emph{ObtainDiamond} task, the agent tries to learn a policy $\pi(a_t|s_t)$ that can maximize $\mathbb{E}[G]$ using limited human demonstrations and environment interactions, where $G=\Sigma^{H}_{t=0}\gamma^{t}r_t$ is the return, and $H$ is the time horizon. Reinforcement learning \cite{sutton2018reinforcement} and imitation learning \cite{osa2018algorithmic} are very general approaches to solve the \emph{ObtainDiamond} MDP problem.

\begin{figure}[t!]
    \centering
    \includegraphics[width=0.98\textwidth]{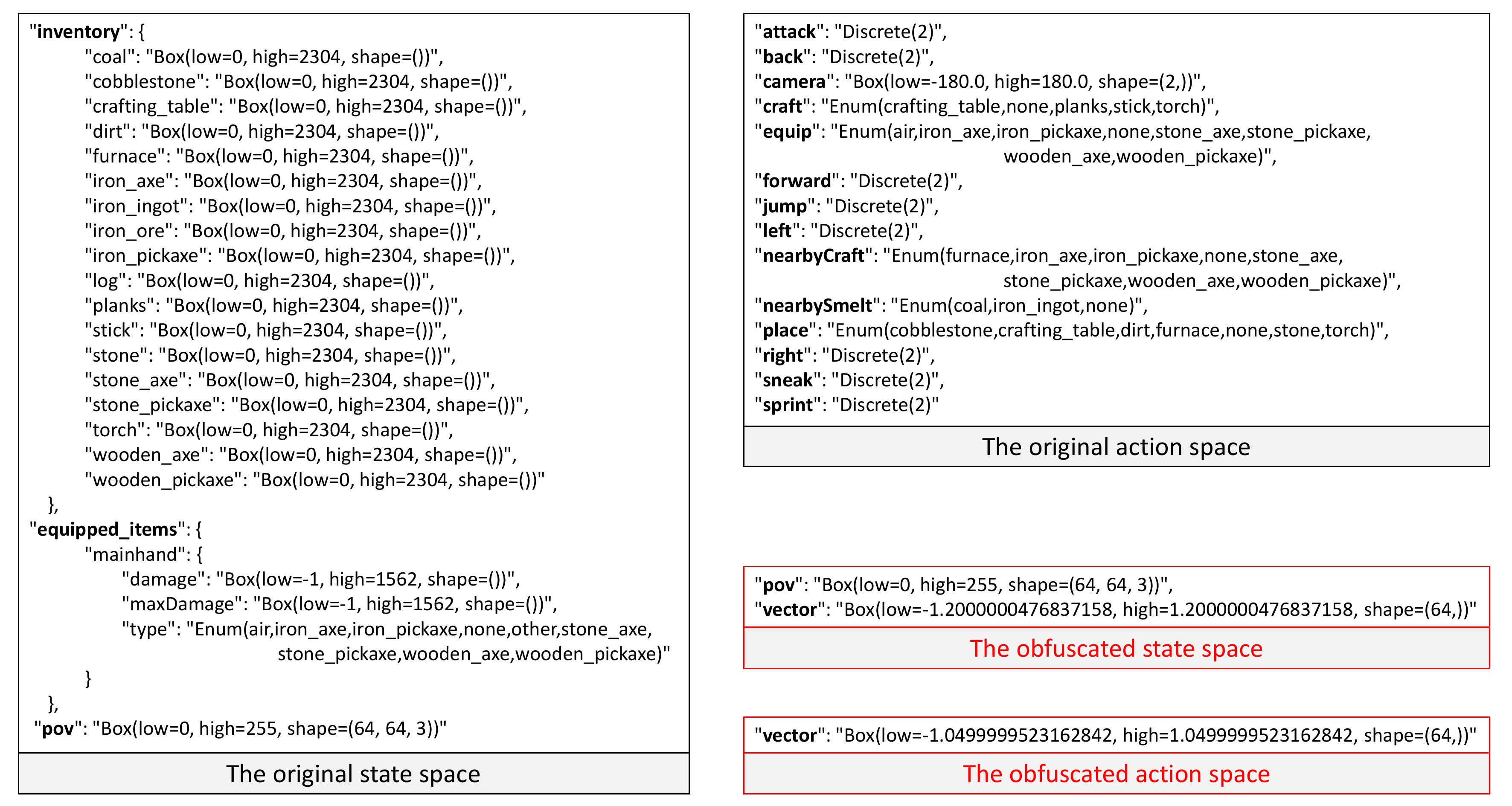}
    \caption{The state and action spaces in the \emph{ObtainDiamond} task. The obfuscated states and actions are more complex since we do not know their semantic meanings.} 
    \label{fig:MDPInfo}
\end{figure}

In the \emph{ObtainDiamond} task, the state and action spaces are very complex. As can be seen from Figure \ref{fig:MDPInfo}, the original state space consists of two parts: the inventory state space and the image observation space. An inventory state includes a number of different items (i.e, the inventory in Figure \ref{fig:MDPInfo}) and the state of the equipped items (i.e, the equipped\_items in Figure \ref{fig:MDPInfo}). An image observation is a game screen shot (i.e, the 64*64*3-dimensional pov in Figure \ref{fig:MDPInfo}). The original action space includes moving left/right/forward/back, attacking, crafting tables, placing tables, rotating the camera, nearby crafting, nearby smelting and other necessary actions. To prevent participants from directly using 
these states and actions which have clear semantic meanings, both the inventory state space and the action space are encoded into 64-dimensional vector spaces, which further increases the difficulty of the competition, as shown in Figure \ref{fig:MDPInfo}. 

\section{Method}
\subsection{The Overall Framework}
\begin{figure}[t!]
    \centering
    \includegraphics[width=0.92\textwidth]{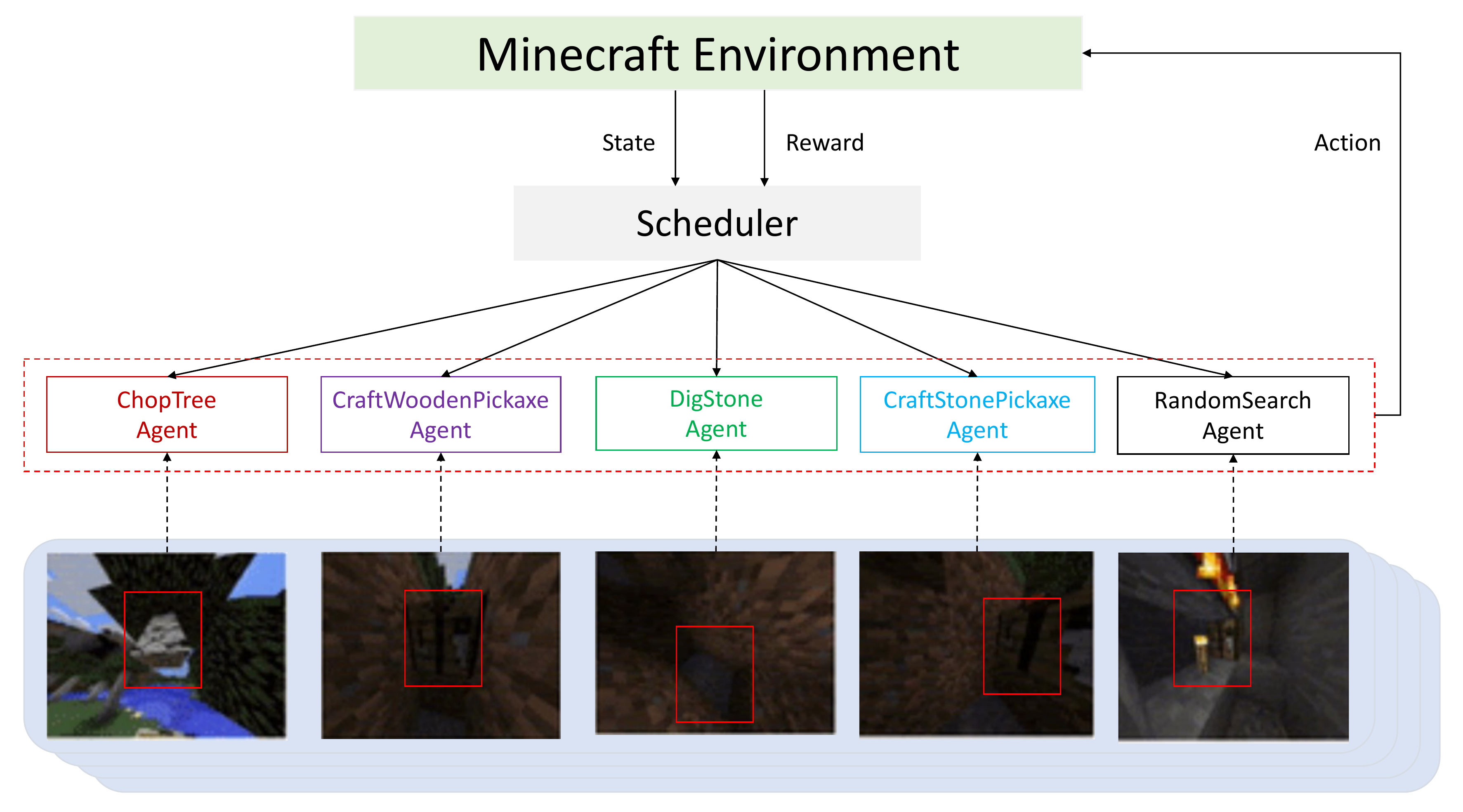}
    \caption{The proposed architecture of our method.} 
    \label{fig:MethodFramework}
\end{figure}

As can be seen from the item hierarchy shown in Figure \ref{fig:ItemHierarchy}, mining a Diamond depends on the completion of a series of different subtasks, each requires mastering different skills. For example, at the beginning of an episode, the agent should search for the Log; after that, the agent should craft the Wooden\_Pickaxe; when this is done, the agent should dig down and search for Cobblestone and Iron\_Ore under the ground. It is hard to imagine how a single policy could learn all these skills (without forgetting other skills). Therefore, we propose to split the \emph{ObtainDiamond} task into five subtasks according to the item dependency, and train a reinforcement learning agent or imitation learning agent (i.e., subpolicies) suitable for each subtask. Specifically, there are five agents: the ChopTree, CraftWoodenPickaxe, DigStone, CraftStonePickaxe, and RandomSearch agent.

A scheduler is needed for choosing which agent to use in different scenario. One way to implement the scheduler is to switch over different agents based on the reward obtained during execution, as obtaining a reward often means the completion of a subtask. However, this method is sometimes problematic and does not generalize well. For example, to craft the Wooden\_Pickaxe, the ChopTree agent should collect at least 3 Logs. Consequently, switching to the CraftWoodenPickaxe agent after the ChopTree agent getting a reward $r=1$ would cause failure of the entire task. Second, it will violate the rules because the competition requires that the submitted method should not include any meta-actions or rule-based heuristics. To make our method totally automatic and general, we propose to implement a learning-based scheduler to select different agents in different states. Specifically, the scheduler is trained using human demonstrations and the supervised learning technique. In the inference phase, it takes as input the current state and predicts which agent to activate. The overall framework of our hierarchical method SEIHAI is illustrated in Figure \ref{fig:MethodFramework}.

In the following subsections, we give a detailed introduction of each agent and the scheduler.

\subsection{Action Discretization} \label{sec:ActionDiscretization}
Because of the domain agnosticism requirement, we can only get the obfuscated states and actions. However, the obfuscated action space is very large, which is a 64-dimensional vector space and each dimension takes a continuous value ranging from $-1.049$ to $1.049$, as shown in Figure \ref{fig:MDPInfo}. As far as we know, current reinforcement learning or imitation learning algorithms can hardly handle this challenge. For example, we conduct preliminary experiments using Deep Deterministic Policy Gradient (DDPG) \cite{silver2014deterministic,lillicrap2016continuous}, an algorithm that is widely used to handle continuous action space, but the performance is unsatisfactory.

To deal with the continuous action space, we transform the MDP with a continuous action space into a MDP with a discrete action space, which can be efficiently solved by algorithms specifically designed for MDPs with discrete action spaces such as Deep Q-network (DQN) \cite{mnih2015human}.

We adopt different techniques to discretize actions in different phases. For the ChopTree agent, the DigStone agent and the RandomSearch agent, we first categorize their corresponding actions into two sets depending on whether an action results in inventory state changes,
 since we observe that the two sets of actions play different roles: generally,  actions that do not incur inventory state changes are responsible for agent movement, such as moving forward and backward; by contrast, actions that contribute to inventory state changes are those for chopping trees and digging stones.   
We leverage the K-Means \cite{macqueen1967some,lloyd1982least,likas2003global} algorithm to cluster each action set into 30 classes. As a result, we transform the continuous action space into a discrete action set with 60 actions in total. For the CraftWoodenPickaxe agent and the CraftStonepickaxe agent, we only extract the actions that contribute to inventory state changes and cluster them into several categories using the Non-parametric Bayesian method \cite{blei2010nested}. We neglect actions that do not incur inventory state changes since crafting different items only requires the inventory information. 

\subsection{The ChopTree Agent}
At the beginning of the \emph{ObtainDiamond} task, the agent is initialized at a random place on a randomly-generated Minecraft map, with nothing in its inventory. According to the item hierarchy shown in Figure \ref{fig:ItemHierarchy}, the primary goal of SEIHAI is to chop trees and collect enough Logs. 

We implement the ChopTree agent using a deep neural network. The network consists of three 2D convolutional layers and two fully connected layers, with a dueling architecture at the last layer to generate the Q-values for every discrete actions. Although this network is very simple, it can balance the training efficiency and testing performance very well\footnote{For example, we test deep ResNet \cite{he2016identity}, but the performance has not improved much, while the training time has increased a lot.}.

We train the agent using the Soft Q Imitation Learning (SQIL) algorithm \cite{reddy2019sqil}. It encourages the agent to follow the trajectories in human demonstrations by giving a penalty (i.e., negative reward) when the agent generates actions that lead to the out-of-distribution states. It has been shown that SQIL can be interpreted as a regularized variant of behavioral cloning that uses a sparse prior to encourage imitation. Therefore, the agent can easily obtain the Logs as human does after training\footnote{We also test the Generative Adversarial Imitation Learning (GAIL) algorithm \cite{ho2016generative}, but it is not as stable as SQIL.}.

Specifically, the training process consists of two-stage. In the first stage, we train the agent using an additional \emph{TreeChop} environment provided by the competition organizer, so that the agent could focus on the chopping tree task. While in the second stage, we fine-tune the agent using the \emph{ObtainDiamond} environment. In the fine-tune phase, we only use the truncated human demonstrations which contain treechop-related interactions only. The trick is crucial to final performance since demonstrations of digging or crafting can mislead the training of the ChopTree agent. To be specific, we only use demonstrations before the reward of crafting a plank is obtained.

In order to deal with the obfuscated high-dimensional observation space, we add an auxiliary task to help with feature extraction. Ideally, an efficient strategy should learn the segmentation of different parts in observations, as well as recognizing their semantic meanings. For instance, it should recognize the position of trees. Therefore, we add an reconstruction loss to the original reinforcement learning objective.

\subsection{The CraftWoodenPickaxe Agent}
In this phase, the agent's goal is to craft a Wooden\_Pickaxe that can be used to obtain other items in the following phase. To this end, the agent needs to first craft Planks using Logs, then make a Crafting\_Table and some Sticks using Planks. Finally, the agent needs to place the Crafting\_Table in an empty area, stand nearby the Crafting\_Table and then craft a Wooden\_Pickaxe. It is worth noting that the Sticks and the Crafting\_Table can be crafted in a random order and the Crafting\_Table can be placed before or after Sticks are crafted. 

Since the behaviors in this phase show a clear pattern, we implement the CraftWoodenPickaxe agent using supervised learning, also known as behavior cloning in the sequential decision making setting. Specifically, we first extract those critical actions $\{a_i\}_{i=0}^N$ resulting in inventory state changes and their corresponding inventory states $\{s_i\}_{i=0}^N$ from the demonstrations. Note that the image states are not included since crafting those items only requires inventory information. The extracted actions $\{a_i\}_{i=0}^N$ are further clustered into $K$ classes using the  non-parametric Bayesian clustering method mentioned in Section \ref{sec:ActionDiscretization}. We build a training set using the inventory states and the corresponding discretized actions and use a fully connected network with two hidden layers each with 64 neurons as the function approximator for the classifier.

\subsection{The DigStone Agent}
The CraftWoodenPickaxe agent is followed by the DigStone agent, which aims at digging enough Cobblestones.

The DigStone agent is implemented using a deep neural network that is exactly the same as that of the ChopTree agent. It can balance the training efficiency and executing performance as mentioned above. 

We train the agent based on a variant of imitation learning method, which we call Large Margin Imitation (LarMI). Specifically, LarMI imitates the actions in the demonstrations with a large margin constraint as follows:
\begin{eqnarray}
	J(Q) = \max_{a \in A, a \neq a_D}[Q(s,a) + T] - Q(s, a_D) \label{equ:LarMI}
\end{eqnarray}
where $Q(s,a)$ represents the ``relative value''\footnote{This is different from the Q-value in reinforcement learning, since here $Q(s,a)$ does not estimate the expected cumulative rewards.} of action $a$ in state $s$, and $a_D$ is the action that the demonstrator took in state $s$, and $T$ is the margin threshold. Equation (\ref{equ:LarMI}) forces the relative values of the other actions to be at least a margin lower than the relative value of the demonstrator’s action. This avoids the overestimation of the unseen actions, and the greedy policy induced by $a^* \leftarrow \arg\max_{a} Q(s,a)$ will easily imitate the demonstrations. Note that our method is very similar to the pretraining stage of the Deep Q-learning from Demonstration (DQfD) algorithm \cite{hester2018deep}. 

The reasons that we do not use DQfD or SQIL are two-fold. First, DQfD and SQIL require interacting with the environment, so they will reduce the sample efficiency. Second, since digging stone is an intermediate subtask, we cannot control the environment to switch to the exact states in the demonstrations, which further prevents us from applying DQfD and SQIL.

\subsection{The CraftStonePickaxe Agent}
In this phase, the agent needs to craft a Stone\_Pickaxe and a Furnace for the purpose of mining Iron\_Ore and crafting Iron\_Ingots. 

Similar to the CraftWoodenPickaxe phase, we first extract critical actions related to this subtask and then cluster them into several classes for action discretization. Different from the CraftWoodenPickaxe, the Stone\_Pickaxe and the Furnace can be crafted in a random order, so we implement the CraftStonePickaxe agent with a policy that randomly selects actions from the clustered action set. The Stone\_Pickaxe and the Furnace can be obtained as long as the critical actions are executed for enough times. The online evaluation results demonstrate that this policy works quite well. 

\subsection{The RandomSearch Agent}
After obtaining the Stone\_Pickaxe, the agent should continue to mine Iron\_Ore, which is the raw material for crafting Iron\_Ingot and Iron\_Pickaxe. The agent can further use the Iron\_Pickaxe to obtain the Diamond. However, the Iron\_Ore may locate at any places underground, which is totally unpredictable. Consequently, the agent must search around for Iron\_Ore. To this end, we combine the DigStone agent and a fully random search agent as the RandomSearch agent to obtain Iron\_Ore. Specifically, at each time step, the RandomSearch agent executes an action either proposed by the DigStone agent or randomly selected from the action set. The randomness of the RandomSearch agent is controlled by an exploration threshold that ranges from zero to one and gradually increases over time. The larger the exploration threshold is, the more random the policy is. 

The Iron\_Ingot, the Iron\_Pickaxe and the Diamond can be obtained by designing agents similar to these already introduced agents. However, we find out that the success rate of obtaining the Iron\_Ore has dropped to around 7\% (illustrated in Table \ref{tab:AgentEvaluation}), which leaves little room for optimization and improvement of subsequent agents. Therefore, we employ the RandomSearch agent to address the remaining tasks by setting the exploration threshold to one, in the hope that it can obtain any left items by luck.  

\subsection{The Scheduler}
The scheduler plays the role of determining which agent to use in different states. The sequential dependency of each agents is shown in Figure \ref{fig:MethodFramework}. To finish the \emph{ObtainDiamond} task, one must finish each subtask sequentially. Deviating from the agent dependency (or task dependency) will cause failure of the task. 

\begin{figure}[t!]
    \includegraphics[width=\textwidth]{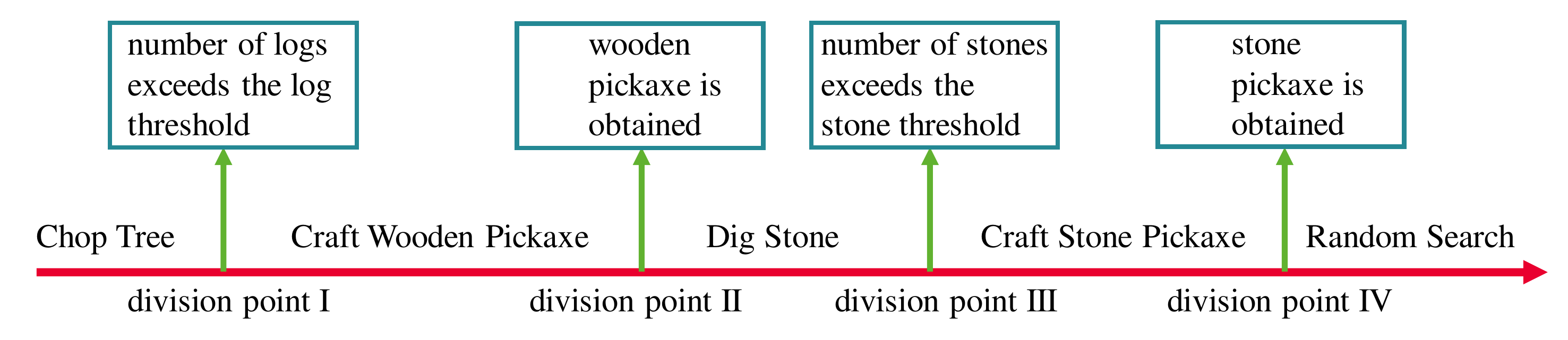}
    \caption{The Division Points of an Episode.} 
    \label{fig:DivisionPoint}
\end{figure}

Given the simple sequential dependency of different agents, training the scheduler boils down to a classification task. To build the training set, we split each episode into five parts, which is detailed in Figure \ref{fig:DivisionPoint}. We set the log threshold as the minimum number\footnote{Note that the minimum number is learned from the demonstrations, and this is allowed by the competition organizers.} of Logs required to finish crafting the Stone\_Pickaxe, and the stone threshold as the minimum number of Cobble\_Stones required to finish crafting the Stone\_Pickaxe and the Furnace. Moreover, we build the training set for the scheduler using the dense demonstrations, since the agent can obtain a certain amount of rewards every time it chops a tree or digs a stone. 

\ignore{
\subsection{Other Implementation Details} \label{sec:CriticalImplementations}
In the Minecraft environment, the state includes the 64*64*3-dimensional pov as shown in Figure \ref{fig:MDPInfo}. Like many deep reinforcement learning algorithms, we adopt several common preprocessings for the pov, including frame skip, frame stack, image graying, pixel value normalization, etc. Besides these common preprocessings, there are some novel implementations as described below.

\subsubsection{Critical Action Extraction.} In order to make our methods more robust and sample efficiency, except for the action discretization mentioned in \ref{sec:ActionDiscretization}, we also ...
}

\section{Experiments}
We present the official results \footnote{https://www.aicrowd.com/challenges/neurips-2020-minerl-competition/leaderboards} reported by the MineRL competition organizers. There are 90+ teams and around 500 submissions in total. The official evaluation consists of two rounds. In round 1, the submitted model is evaluated. In round 2, only codes are submitted, and the model is retrained by the competition organizer from scratch. The final score is averaged over 200 episodes. For each episode, the score is computed as the sum of the rewards (shown in Figure \ref{fig:ItemHierarchy}) achieved by the agent. 

\subsection{Overall Evaluation}
Table \ref{tab:OverallEvaluation} shows the scores of the best-performing submissions from both rounds. As can be seen, our methods rank first in the preliminary and final of the NeurIPS-2020 competition. Besides, scores obtained by our methods surpass those of the second-place team by a large margin, demonstrating the effectiveness of our methods. It is also worth noting that our method can be further improved: our round 1 submission only consists of the ChopTree, the CraftWoodenPickaxe, and the DigStone agents, and the corresponding score is 19.84; in contrast, we further add the CraftStonePickaxe and the RandomSearch agents in round 2, and the score becomes 39.55. If we design specific agents to craft Iron\_Ingot and Iron\_Pickaxe, we believe the score can be further improved. We leave it as our future work.

\begin{table}[t!]
\centering
\caption{The final scores of the official baselines (left) and the best-performing submissions from Round 1 (middle)
and Round 2 (right). Note that only six teams achieved a non-zero score in round 2. Our team is HelloWorld.}\label{tab:OverallEvaluation}  
\setlength{\tabcolsep}{2.5mm}{
    \begin{tabular}{|c|r|c|r|c|r|}
    \hline
    \multicolumn{2}{|c|}{\bfseries Baselines} & \multicolumn{2}{c|}{\bfseries Round 1 (Preliminary)} & \multicolumn{2}{c|}{\bfseries Round 2 (Final)} \\
    \hline
    \bfseries Name & \bfseries Score & \bfseries Team Name & \bfseries Score & \bfseries Team Name & \bfseries Score \\
    \hline
    SQIL & 2.94 & \bfseries HelloWorld & \bfseries 19.84 & \bfseries HelloWorld & \bfseries 39.55 \\
    DQfD & 2.39 & NoActionWasted & 16.48 &  michal\_opanowicz & 13.29 \\
    Rainbow & 0.42 & michal\_opanowicz & 9.29 & NoActionWasted & 12.79 \\
    PDDDQN & 0.11 & CU-SF & 6.47 & Rabbits & 5.16 \\
     &  & NuclearWeapon & 4.34 & MajiManji & 2.49 \\
     &  & murarinCraft & 3.61 & BeepBoop & 1.97 \\
    \hline
    \end{tabular}
}
\end{table}

\subsection{Agent-level Evaluation}

\begin{table}[t!]
\caption{Details of our final submission. The ``\# Episode'' column shows the total number of episodes end up with the corresponding score. The ``Cond. Rate'' column depicts the conditional success rate of each agent given the success of its previous agent. The ``Rate'' column shows the success rate of SEIHAI in different phases.}\label{tab:AgentEvaluation}
\centering
    \begin{tabular}{c|c|c|c|c|c}
    \hline
    \bfseries Agent & \bfseries Item & \bfseries Score & \bfseries \# Episode & \bfseries Cond. Rate & \bfseries Rate \\
    \hline
    \multirow{2}*{\bfseries ChopTree}  & None & 0 & 72 & \multirow{2}*{64.0\%} & \multirow{2}*{64.0\%} \\
    & Log & 1 & 11 &  & \\
    \hline
    \multirow{4}*{\bfseries CraftWoodenPickaxe} & Plank & 3& 15 & \multirow{4}*{78.6\%} & \multirow{4}*{46.0\%} \\
    & Stick & 7 & 3 &  &  \\
    & Crafting\_Table & 11 & 6 &  &  \\
    & Wooden\_Pickaxe & 19 & 20 &  &  \\
    \hline
    \bfseries DigStone & Cobblestone & 35 & 9 & 78.3\% & 36.0\% \\
    \hline
    \multirow{2}*{\bfseries CraftStonePickaxe} & Stone\_Pickaxe & 67 & 2 & \multirow{2}*{84.7\%} & \multirow{2}*{31.5\%} \\
    & Furnace & 99 & 47 &  & \\
    \hline
    \multirow{4}*{\bfseries RandomSearch} & Iron\_Ore & 163 & 14 & \multirow{4}*{23.0\%} & \multirow{4}*{7.0\%} \\
    & Iron\_Ignots & 291 & 0 &  & \\
    & Iron\_Pickaxe & 547  & 0 &  &  \\
    & Diamond & 1571 & 0 &  &  \\
    \hline
    \end{tabular}
\end{table}

\ignore{
\begin{table}
\caption{Table captions should be placed above the
tables.}\label{tab1}
\centering
\setlength{\tabcolsep}{2mm}{
    \begin{tabular}{|c|c|c|c|c|c|c|c|c|c|c|c|c|}
    \hline
    Sum of Rewards & 0 & 1 & 3 & 7 & 11 & 19 & 35 & 67 & 99 & 163 & & Score \\
    \hline
    \# Episode & 72 & 11 & 15 & 3 & 6 & 20 & 9 & 2 & 47 & 14 & & 39.55 \\
    \hline
    \end{tabular}
}
\end{table}

\begin{table}[t!]
\caption{The score distribution of our best submission (i.e., average score = 39.55).}\label{tab:AgentEvaluation}
\centering
    \begin{tabular}{|c|c|c|c|c|c|}
    \hline
    Score & 0 & 1 & 3 & 7 & 11 \\
    \hline
    Milestone & Empty Inventory & Log & Plank & Stick & Crafting\_Table \\
    \hline
    \# Episode & 72 & 11 & 15 & 3 & 6 \\
    \hline
    \hline
    Sum of Rewards & 19 & 35 & 67 & 99 & 163 \\
    \hline
    Milestone & Wooden\_Pickaxe & Cobbleston & Furnace & Stone\_Pickaxe & Iron\_Ore \\
    \hline
    \# Episode & 20 & 9 & 2 & 47 & 14 \\
    \hline
    \end{tabular}
\end{table}

\begin{table}[t!]
\caption{The success rate of different agents in our best submission. Agent1, agent2, agent3, agent4 and agent5 represent the ChopTree, CraftWoodenPickaxe, DigCobblestone, CraftStonePickaxe, and RandomSearch agent, respectively.}\label{tab:SuccessRateEvaluation}
\centering
\setlength{\tabcolsep}{2.5mm}{
    \begin{tabular}{|c|c|c|c|c|c|}
    \hline
    Agent & Agent1 & Agent2 & Agent3 & Agent4 & Agent5 \\
    \hline
    Success Rate & 64.00\% & 46.50\% & Plank & Stick & Crafting\_Table \\
    \hline
    \end{tabular}
}
\end{table}
}

Table \ref{tab:AgentEvaluation} demonstrates the detailed performance of our final submission, which is evaluated for 200 episodes. We can see that there are 72 episodes
with zero rewards, meaning that the ChopTree agent fails to collect any Logs
in the online evaluation environment. The reason is that in the ObtainDiamond environment, the agent often starts off from some deserted places (e.g., in water, in the desert or on the cliff). As a result, the ChopTree agent must search around for trees, which may be risky or take a lot of time. We believe that the ChopTree agent can be further improved by developing techniques handling such extreme cases. Fortunately, the CraftWoodenPickaxe agent, the DigStone agent and the Craft-
StonePickaxe agent perform well, achieving a conditional success rate of more
than 78.0\%. However, as different agents are called in tandem, the success rate of SEIHAI in a certain
phase is the multiplication of the conditional success rates of all the previously
activated agents and would degrades exponentially as more agents are called. Evaluation results show that the success rate of SEIHAI at the crafting Stone\_Pickaxe phase has dropped to 31.5\%, which is very low.

The RandomSearch agent is responsible for the remaining tasks after obtaining the Stone\_Pickaxe. Results show that this policy can obtain the Iron\_Ore with a conditional success rate of 23.0\%, and the success rate of SEIHAI in this phase drops to 7.0\%. Since we do not design specific agents to attack these tasks due to the complexity of the underground environment and the difficulty of obtaining effective actions, we believe that well-designed agents for obtaining Iron\_Ore and subsequent items can further improve the overall performance.

\section{Related Work}
\textbf{The MineRL competition} \cite{guss2019minerldataset,guss2019neurips,milani2020retrospective,guss2021towards,guss2021minerl,guss2019minerl} has become a popular competition promoting sample-efficient reinforcement learning and imitation learning algorithms. There are lots of relevant studies. For example, HDQfD \cite{skrynnik2021hierarchical} utilizes the hierarchical structure of expert trajectories, presenting a structured task-dependent replay buffer and an adaptive prioritizing technique to gradually erase poor-quality expert data from the buffer. HDQfD won the first place in 2019. The runner-up method \cite{amiranashvili2020scaling} highlights the influence of network architecture, loss function, and data augmentation. In contrast, the 
third winner \cite{scheller2020sample} proposes a training procedure where policy networks are first trained with human demonstrations and later fine-tuned by reinforcement learning; additionally, they propose a policy exploitation mechanism, an experience replay scheme and an additional loss regularizer to prevent catastrophic forgetting of previously learned skills. However, these methods include many hand-crafted meta-actions or rules, which cannot generalize to new environment and is the main reason that the MineRL 2020 introduces the domain agnosticism requirement. Different from these methods, our method is domain-agnostic and generalizes well in new online training and evaluation environments.

\textbf{Learning from demonstration} is an effective way to achieving data efficiency for the MineRL competition. The simplest way is behaviour cloning, but it cannot handle complex situations. Recent studies such as DQfD \cite{hester2018deep}, DDPGfD \cite{vecerik2017leveraging}, POfD \cite{kang2018policy}, SQIL \cite{reddy2019sqil} and GIAL \cite{ho2016generative} are more effective, but they can hardly handle the challenging MineRL environment with very sparse rewards. Consequently, we only adopt some of them to attack subtasks in our work. We notice that offline reinforcement learning algorithms like BCQ \cite{fujimoto2019off}, BEAR \cite{kumar2019stabilizing} and CQL \cite{kumar2020conservative} are also promising approaches, and we leave the exploration of these methods as our future work.

\section{Conclusion}
In this work we present a sample-efficient hierarchical method to handle the \emph{ObtainDiamond} task with limited human demonstrations, sparse rewards but an explicit task structure. The task is divided into several sequentially dependent subtasks (e.g., Treechop task, CraftWoodenPickaxe task, and DigCobblestone task) based on human priors. Furthermore, we design an imitation learning or a reinforcement learning agent suitable for each subtask. Besides, an imitation learning based scheduler is developed to determine which agent to use in different states. We win the first place in the preliminary and final of the NeurIPS-2020 MineRL competition, which demonstrates the efficiency of our hierarchical method, SEIHAI. We also identify several directions for further improvement based on detailed analysis of the online evaluation results. We believe that developing methods that properly combine human priors and sample-efficient learning-based techniques is a competitive way to solve complex tasks with limited demonstrations, sparse rewards but an explicit task structure.

\section*{Acknowledgement}
The authors would like to thank Mengchen Zhao, Weixun Wang, Rundong Wang, Shixun Wu, Zhanbo Feng and the anonymous
reviewers for their comments.

\bibliographystyle{splncs04}
\bibliography{mybibliography}

\begin{thebibliography}{10}
\providecommand{\url}[1]{\texttt{#1}}
\providecommand{\urlprefix}{URL }
\providecommand{\doi}[1]{https://doi.org/#1}

\bibitem{amiranashvili2020scaling}
Amiranashvili, A., Dorka, N., Burgard, W., Koltun, V., Brox, T.: Scaling
  imitation learning in minecraft. arXiv preprint arXiv:2007.02701  (2020)

\bibitem{blei2010nested}
Blei, D.M., Griffiths, T.L., Jordan, M.I.: The nested chinese restaurant
  process and bayesian nonparametric inference of topic hierarchies. Journal of
  the ACM (JACM)  \textbf{57}(2),  1--30 (2010)

\bibitem{fujimoto2019off}
Fujimoto, S., Meger, D., Precup, D.: Off-policy deep reinforcement learning
  without exploration. In: International Conference on Machine Learning. pp.
  2052--2062. PMLR (2019)

\bibitem{gu2017deep}
Gu, S., Holly, E., Lillicrap, T., Levine, S.: Deep reinforcement learning for
  robotic manipulation with asynchronous off-policy updates. In: 2017 IEEE
  international conference on robotics and automation (ICRA). pp. 3389--3396.
  IEEE (2017)

\bibitem{guss2021minerl}
Guss, W.H., Castro, M.Y., Devlin, S., Houghton, B., Kuno, N.S., Loomis, C.,
  Milani, S., Mohanty, S., Nakata, K., Salakhutdinov, R., et~al.: The minerl
  2020 competition on sample efficient reinforcement learning using human
  priors. arXiv preprint arXiv:2101.11071  (2021)

\bibitem{guss2019neurips}
Guss, W.H., Codel, C., Hofmann, K., Houghton, B., Kuno, N., Milani, S.,
  Mohanty, S., Liebana, D.P., Salakhutdinov, R., Topin, N., et~al.: Neurips
  2019 competition: The minerl competition on sample efficient reinforcement
  learning using human priors. arXiv preprint arXiv:1904.10079  (2019)

\bibitem{guss2019minerl}
Guss, W.H., Codel, C., Hofmann, K., Houghton, B., Kuno, N., Milani, S.,
  Mohanty, S., Perez~Liebana, D., Salakhutdinov, R., Topin, N., et~al.: The
  minerl competition on sample efficient reinforcement learning using human
  priors. arXiv e-prints  (2019)

\bibitem{guss2019minerldataset}
Guss, W.H., Houghton, B., Topin, N., Wang, P., Codel, C., Veloso, M.,
  Salakhutdinov, R.: Minerl: A large-scale dataset of minecraft demonstrations.
  arXiv preprint arXiv:1907.13440  (2019)

\bibitem{guss2021towards}
Guss, W.H., Milani, S., Topin, N., Houghton, B., Mohanty, S., Melnik, A.,
  Harter, A., Buschmaas, B., Jaster, B., Berganski, C., et~al.: Towards robust
  and domain agnostic reinforcement learning competitions: Minerl 2020. In:
  NeurIPS 2020 Competition and Demonstration Track. pp. 233--252. PMLR (2021)

\bibitem{he2016identity}
He, K., Zhang, X., Ren, S., Sun, J.: Identity mappings in deep residual
  networks. In: European conference on computer vision. pp. 630--645. Springer
  (2016)

\bibitem{hester2018deep}
Hester, T., Vecerik, M., Pietquin, O., Lanctot, M., Schaul, T., Piot, B.,
  Horgan, D., Quan, J., Sendonaris, A., Osband, I., et~al.: Deep q-learning
  from demonstrations. In: Thirty-second AAAI conference on artificial
  intelligence (2018)

\bibitem{ho2016generative}
Ho, J., Ermon, S.: Generative adversarial imitation learning. Advances in
  neural information processing systems  \textbf{29},  4565--4573 (2016)

\bibitem{kang2018policy}
Kang, B., Jie, Z., Feng, J.: Policy optimization with demonstrations. In:
  International Conference on Machine Learning. pp. 2469--2478. PMLR (2018)

\bibitem{kumar2019stabilizing}
Kumar, A., Fu, J., Tucker, G., Levine, S.: Stabilizing off-policy q-learning
  via bootstrapping error reduction. arXiv preprint arXiv:1906.00949  (2019)

\bibitem{kumar2020conservative}
Kumar, A., Zhou, A., Tucker, G., Levine, S.: Conservative q-learning for
  offline reinforcement learning. arXiv preprint arXiv:2006.04779  (2020)

\bibitem{likas2003global}
Likas, A., Vlassis, N., Verbeek, J.J.: The global k-means clustering algorithm.
  Pattern recognition  \textbf{36}(2),  451--461 (2003)

\bibitem{lillicrap2016continuous}
Lillicrap, T.P., Hunt, J.J., Pritzel, A., Heess, N., Erez, T., Tassa, Y.,
  Silver, D., Wierstra, D.: Continuous control with deep reinforcement
  learning. In: ICLR (2016)

\bibitem{lloyd1982least}
Lloyd, S.: Least squares quantization in pcm. IEEE transactions on information
  theory  \textbf{28}(2),  129--137 (1982)

\bibitem{macqueen1967some}
MacQueen, J., et~al.: Some methods for classification and analysis of
  multivariate observations. In: Proceedings of the fifth Berkeley symposium on
  mathematical statistics and probability. vol.~1, pp. 281--297. Oakland, CA,
  USA (1967)

\bibitem{mao2017accnet}
Mao, H., Gong, Z., Ni, Y., Xiao, Z.: Accnet: Actor-coordinator-critic net for"
  learning-to-communicate" with deep multi-agent reinforcement learning. arXiv
  preprint arXiv:1706.03235  (2017)

\bibitem{mao2020neighborhood}
Mao, H., Liu, W., Hao, J., Luo, J., Li, D., Zhang, Z., Wang, J., Xiao, Z.:
  Neighborhood cognition consistent multi-agent reinforcement learning. In:
  Proceedings of the AAAI Conference on Artificial Intelligence. vol.~34, pp.
  7219--7226 (2020)

\bibitem{mao2019modelling}
Mao, H., Zhang, Z., Xiao, Z., Gong, Z.: Modelling the dynamic joint policy of
  teammates with attention multi-agent ddpg. In: Proceedings of the 18th
  International Conference on Autonomous Agents and MultiAgent Systems (2019)

\bibitem{mao2020learningAgent}
Mao, H., Zhang, Z., Xiao, Z., Gong, Z., Ni, Y.: Learning agent communication
  under limited bandwidth by message pruning. In: Proceedings of the AAAI
  Conference on Artificial Intelligence. vol.~34, pp. 5142--5149 (2020)

\bibitem{mao2020learningMultiAgent}
Mao, H., Zhang, Z., Xiao, Z., Gong, Z., Ni, Y.: Learning multi-agent
  communication with double attentional deep reinforcement learning. Autonomous
  Agents and Multi-Agent Systems  \textbf{34}(1),  1--34 (2020)

\bibitem{milani2020retrospective}
Milani, S., Topin, N., Houghton, B., Guss, W.H., Mohanty, S.P., Nakata, K.,
  Vinyals, O., Kuno, N.S.: Retrospective analysis of the 2019 minerl
  competition on sample efficient reinforcement learning. In: NeurIPS 2019
  Competition and Demonstration Track. pp. 203--214. PMLR (2020)

\bibitem{mnih2015human}
Mnih, V., Kavukcuoglu, K., Silver, D., Rusu, A.A., Veness, J., Bellemare, M.G.,
  Graves, A., Riedmiller, M., Fidjeland, A.K., Ostrovski, G., et~al.:
  Human-level control through deep reinforcement learning. nature
  \textbf{518}(7540),  529--533 (2015)

\bibitem{osa2018algorithmic}
Osa, T., Pajarinen, J., Neumann, G., Bagnell, J.A., Abbeel, P., Peters, J.,
  et~al.: An algorithmic perspective on imitation learning. Foundations and
  Trends in Robotics  \textbf{7}(1-2),  1--179 (2018)

\bibitem{reddy2019sqil}
Reddy, S., Dragan, A.D., Levine, S.: Sqil: Imitation learning via reinforcement
  learning with sparse rewards. In: ICLR (2019)

\bibitem{scheller2020sample}
Scheller, C., Schraner, Y., Vogel, M.: Sample efficient reinforcement learning
  through learning from demonstrations in minecraft. In: NeurIPS 2019
  Competition and Demonstration Track. pp. 67--76. PMLR (2020)

\bibitem{silver2014deterministic}
Silver, D., Lever, G., Heess, N., Degris, T., Wierstra, D., Riedmiller, M.:
  Deterministic policy gradient algorithms. In: ICML. pp. 387--395. PMLR (2014)

\bibitem{skrynnik2021hierarchical}
Skrynnik, A., Staroverov, A., Aitygulov, E., Aksenov, K., Davydov, V., Panov,
  A.I.: Hierarchical deep q-network from imperfect demonstrations in minecraft.
  Cognitive Systems Research  \textbf{65},  74--78 (2021)

\bibitem{sutton2018reinforcement}
Sutton, R.S., Barto, A.G.: Reinforcement learning: An introduction. MIT press
  (2018)

\bibitem{vecerik2017leveraging}
Vecerik, M., Hester, T., Scholz, J., Wang, F., Pietquin, O., Piot, B., Heess,
  N., Roth{\"o}rl, T., Lampe, T., Riedmiller, M.: Leveraging demonstrations for
  deep reinforcement learning on robotics problems with sparse rewards. arXiv
  preprint arXiv:1707.08817  (2017)

\end{thebibliography}

\end{document}